\documentclass[conference]{IEEEtran}
\IEEEoverridecommandlockouts
\usepackage{cite}
\usepackage{amsmath,amssymb,amsfonts}
\usepackage{algorithmic}
\usepackage{graphicx}
\usepackage{textcomp}
\usepackage{tikz}
\usepackage{xcolor}
\usepackage{url}
\usepackage{subcaption}
\def\BibTeX{{\rm B\kern-.05em{\sc i\kern-.025em b}\kern-.08em
    T\kern-.1667em\lower.7ex\hbox{E}\kern-.125emX}}
\begin{document}

\title{Active hypothesis testing in unknown environments using recurrent neural networks and model free reinforcement learning
}

\author{\IEEEauthorblockN{ George Stamatelis}
\IEEEauthorblockA{\textit{Department of Informatics and Telecommunications} \\
\textit{National and Kapodistrian University of Athens}\\
Athens, Greece \\
georgestamat at di.uoa.gr}
\and
\IEEEauthorblockN{ Nicholas Kalouptsidis}
\IEEEauthorblockA{\textit{Department of Informatics and Telecommunications} \\
\textit{National and Kapodistrian University of Athens}\\
Athens, Greece \\
kalou at di.uoa.gr}

}

\maketitle

\begin{abstract}
A combination of deep reinforcement learning and supervised learning is proposed for active sequential hypothesis testing in completely unknown environments. We make no assumptions about the prior probability, the action and observation sets, as well as the observation generating process. Experiments with synthetic Bernoulli and real cybersecurity data demonstrate that our method  performs competitively and sometimes better than the Chernoff test, in both finite and infinite horizon problems. 
\end{abstract}

\begin{IEEEkeywords}
Active hypothesis testing, POMDPs, Reinforcement learning, Recurrent Neural networks, Controlled sensing, Supervised learning, Sensor networks, Anomaly detection
\end{IEEEkeywords}

\section{Introduction}
In active sequential hypothesis testing (ASHT),  a decision maker actively conducts a sequence of experiments in order to infer the underlying hypothesis from a finite set of hypotheses. The problem finds numerous applications including anomaly detection \cite{activeHTforAD}, medical diagnosis \cite{Berry2010BayesianAM}, radar assisted target search
\cite{chernoff-radar} and content search applications \cite{fabioAdaptiveAHT}. It has been  extended to environments involving passive eavesdroppers \cite{evasive-active} or active adversaries \cite{aaht}. 
Deep reinforcement learning (DRL) techniques such as DQN  \cite{mnih2015humanlevel} and deep actor critic algorithms were recently applied to ASHT (see \cite{KatrikEtAll}, \cite{ActorCriticHT} and \cite{aaht}).  In these studies, the DRL agent selects  actions based on beliefs over all possible hypotheses. 
Multiagent DRL methods for anomaly detection were proposed in \cite{decHTKohen} and \cite{decADRL}.

In this work we consider belief free methods. The states of Deep recurrent neural networks (RNN)s are used as proxies for beliefs. We combine  RNNs with reinforcement learning (RL)  to address ASHT in environments with unknown observation probabilities and possibly infinite or continuous observations. At each time instance, the decision maker selects which experiment to conduct based on the history of past observations and actions. This is performed by a RNN network called RNNpolicy.  Once the episode is terminated, the inference network, termed RNNinference  is used to infer the hypothesis. In the infinite horizon setting, an additional RNN network, the so called RNNmonitor monitors the decision maker's interaction with the environment and determines whether the experiments should terminate or continue. We assume that a training environment or a large dataset is available to train the networks. The actual environment in which the networks will be deployed can be slightly different, and the observation probabilities are unknown. Therefore, the agent can not take advantage of the belief vectors. In the case of anomaly detection over sensor networks \cite{activeHTforAD}, the size of the belief vector increases exponentially with the number of sensors. Consequently, quick decision making on large sensor networks by recursively updating beliefs  and  passing it through a large neural network to select the next query, may prove infeasible.

We apply a training procedure that includes the following steps. First we train the RNNpolicy on an artificial simulation environment using a DRL algorithm. A large dataset of actions and observations is created using the RNNpolicy network. Then the RNNmonitor is trained on  data generated by the RNNpolicy using backpropagation \cite{backpropagation}. Finally the pair RNNpolicy-RNNmonitor is used to create a second large dataset and the RNNinference is trained using backpropagation. 

\subsection{Related work}
In his seminal work \cite{ChernoffHT}, Chernoff proposed a heuristic called the Chernoff test and proved that it is asymptotically optimal under certain assumptions. In \cite{controlledSensForMH}, a new variant of the test was introduced in the general multi-hypothesis testing. In \cite{Naghshvar_2013} ASHT is modeled as a cost minimisation  POMDP. The cost depends on the stopping time and the error probability. In \cite{KatrikEtAll}, ASHT is modeled as a confidence maximisation belief MDP and a recurrent DQN  \cite{mnih2015humanlevel} procedure is proposed and numerically shown to perform better than other heuristics. Similarly, in \cite{ActorCriticHT} a deep actor critic procedure is shown to perform better than the Chernoff test in environments that require high confidence levels.

\subsection{Our contribution}
We propose a combination of DRL algorithms and RNNs to  solve ASHT problems in completely unknown environments. We make no assumptions about the prior probability of the hypotheses, the  observation generating process or the type of observations. In fact continuous action spaces are allowed. The only requirement is access to a training environment, or an appropriate training dataset.    We compare our approach to the modified Chernoff test of \cite{controlledSensForMH} in environments with discrete, finite actions and observations, and show that the proposed method achieves competitive results and sometimes performs slightly better.

The rest of this paper is organised as follows. ASHT  is defined in section II. The proposed method is presented in section III  for finite and infinite horizons. Comparisons with the Chernoff test are given in sections IV and V using  anomaly detection environments. Further information about the experiments and sample efficiency analysis is presented as appendix in the extended version.

\section{Problem statement}
Let $\mathcal{X}=\{0,1,...,N\}$ be a finite set of hypotheses. The random variable $X$ defines the true hypothesis. At each time $t>0$ a decision maker chooses an action $a_t \in \mathcal{A}$. Based on $a_t$ and $X=x$, an observation $y_t \in \mathcal{Y}$ is generated. Typically it is assumed that these observations are i.i.d, sampled from a known distribution 
\begin{math}
    \label{obsProb}
   p_X^{a_t} \triangleq P[Y|X,a_t],
\end{math}
while both actions and observations belong to finite sets. None of these assumptions are required in this work. 
Actions aim to assist inference of the true hypothesis. They rely on 
the information available at time $t$:
\begin{math}
    \label{infoSet}
    I_t=\{a_{1:t-1},y_{1:t-1}\},
\end{math}
and  are sampled from a belief based policy.
For a given action $a_t$ and observation $y_t$, the belief on each hypothesis $i \in \mathcal{X}$ is updated recursively as \begin{equation}
    \label{belup}
    \rho_{t+1}(i)= \rho_t(i) \frac{p_i^{a_t}(y_t)}{\sum_j \rho_t(j) p_j^{a_t}(y_t)}.
\end{equation}
$\rho_0$ is the prior probability that a hypothesis is true.

The inferred hypothesis is denoted by $\hat {X}_t$ and is the result of the inference policy acting on the information set. 

The quality of inference is measured by several indicators. The error probability is expressed in terms of beliefs as  
\begin{math}
 \gamma_t=P[\hat{X}_t\neq X] = 1-\max_{i}\rho_t(i).
\end{math}
The average confidence level is given by
\begin{equation}
    \label{confDef}
    C(\rho)=\sum_{i \in \mathcal{X}} \rho(i) \log \bigg(\frac{\rho(i)}{1-\rho(i)}\bigg).
\end{equation}
 The following index 
\begin{equation}
    \label{LLRatio}
    LL_t=log \frac{p_{\hat{i}_t}(y_{1:t},a_{1:t})}{\max_{j \neq \hat{i}_t} p_j(y_{1:t},a_{1:t})},
\end{equation}
where $\hat{i}_t$ is the most likely hypothesis monitors performance in the infinite horizon setting. It provides an asymptotically optimal stopping rule: the agent terminates the experiment the first time $t$ for which
\begin{math}
    \label{chernoffStop}
    LL_t > -\log c,
\end{math}
where $c$ is a positive real valued parameter.
Upon termination, the inference strategy selects the hypothesis $\hat{i}_t$ that maximizes the aposteriori probability (MAP decoding) or the likelihood (ML decoding). Often a uniform prior is employed, in which case the MAP and ML decoding rules coincide. For further details please see \cite{KatrikEtAll} and \cite{aaht}.

An asympotically optimal policy is given by the Chernoff test. At each time $t$, actions are  sampled from the distribution 
\begin{equation}
    \label{chernoffTest}
    g_t = \max_{g} \min_{j \neq \hat{i}_t} \sum_a g(a) D(p_{\hat{i}}^a||p_j^a).
\end{equation}

\section{Architecture and training procedure}
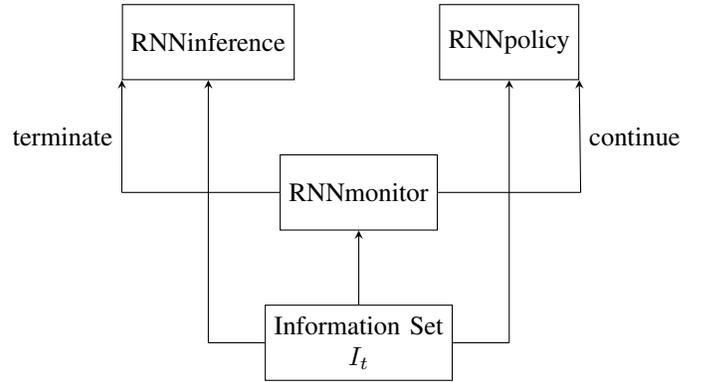
\begin{figure}[]
\centering
\begin{tikzpicture}

\node[draw,
    minimum width=1cm,
    minimum height=1cm,
    align=center
] (It) at (0,0){Information Set \\  $I_t$};
\node[draw,
minimum width=1cm,
    minimum height=1cm,
    align=center
](RNNMonitor) at (0,2) {RNNmonitor};

\node[draw,
minimum width=1cm,
    minimum height=1cm,
    align=center
](RNNInference) at (-2,4) {RNNinference};

\node[draw,
minimum width=1cm,
    minimum height=1cm,
    align=center
](RNNPolicy) at (2,4) {RNNpolicy};

\node[](temp1) at (-2,0){};
\node[](temp2) at (2,0){};

\node[](temp3) at (-3.15,2){};
\node[](temp4) at (2.95,2){};
\draw[-](It)--(temp1.center){};
\draw[-stealth](temp1.center)--
(RNNInference.south){};

\draw[-](It)--(temp2.center){};
\draw[-stealth](temp2.center)--(RNNPolicy.south){};

\draw[-](RNNMonitor)--(temp4.center){};
\draw[-stealth](temp4.center)--(RNNPolicy.south east)node[midway,right]{continue};

\draw[-](RNNMonitor)--(temp3.center){};
\draw[-stealth](temp3.center)--(RNNInference.south west)node[midway,left]{terminate};

\draw[-stealth](It)--(RNNMonitor);
\end{tikzpicture}
\caption{\textbf{Architecture.} RNNmonitor decides whether to continue based on the information set. Then either RNNinference guesses the underlying hypothesis or RNNpolicy continues with experiment selection. } \label{fig:M1}
\end{figure}
In this section, we combine DRL and supervised learning to tackle ASHT without the use of belief vectors. We assume that we have access to a training environment or alternatively, to a large dataset, based on which we can build a simulator.  A schematic plot of the proposed architecture is given in fig \ref{fig:M1}.

Two supervised neural networks are in charge of termination time and inference, and one (or two for actor critic algorithms) is in charge of experiment selection. 
We employ recurrent neural networks (RNN) rather than feedforward neural networks to take advantage of their capacity to model dynamic input output systems. Simple RNNs are prone to numerical instability problems, most notably vanishing and exploding gradients. To remedy these issues, several variations have been proposed, most notably, long short term memory networks (LSTM)s \cite{lstm} and Gated recurrent units (GRU)s \cite{gru}, which have been successfully applied to POMDP agents, \cite{recurrentDQNStone} and \cite{recurrentModelFreeCan}. In this work we consider both LSTM and GRU networks. 

Under the infinite horizon setting we employ two recurrent neural networks for decoding, RNNmonitor and RNNinference. The first network monitors the environment at each time instance. The value determined by its  output is used to decide whether the process continues or terminates on the basis of a threshold rule. The second network is activated when the RNNmonitor decides termination. It produces an estimate of the underlying true hypothesis. 
Under the fixed finite horizon case, the RNNmonitor becomes redundant. The structure of the two recurrent networks is identical except from the final layer which performs the regression and classificatin tasks respectively. 

\begin{figure}[]
\centering
\begin{tikzpicture}

\node[draw,
    minimum width=1cm,
    minimum height=1cm,
    align=center
] (firstIn) at (-5,0){$(a_1,y_1)$};
\node[draw,
minimum width=1cm,
    minimum height=1cm,
    align=center
](secIn) at (-3,0) {$(a_2,y_2)$};
\node[
](thirdin) at (-1,0) {$\cdots$};
\node[draw,
minimum width=1cm,
    minimum height=1cm,
    align=center
](lastIn) at (1,0) {$(a_T,y_T)$};

\node[draw,
    minimum width=1cm,
    minimum height=1cm,
    align=center
] (firstRNN) at (-5,2){RNN};
\node[draw,
minimum width=1cm,
    minimum height=1cm,
    align=center
](secRNN) at (-3,2) {RNN};
\node[
](thirdRNN) at (-1,2) {$\cdots$};
\node[draw,
minimum width=1cm,
    minimum height=1cm,
    align=center
](lastRNN) at (1,2) {RNN};
\draw[-stealth](firstIn)--(firstRNN.south){};
\draw[-stealth](secIn)--(secRNN.south){};
\draw[-stealth](lastIn)--(lastRNN.south){};

\node[draw,
minimum width=1cm,
    minimum height=1cm,
    align=center
](avg) at (-1.75,4){Average};
\draw[-stealth](firstRNN.north)--(avg.south west)node[midway,above]{$o_1$};
\draw[-stealth](secRNN.north)--(avg.south)node[midway,above]{$o_2$};
\draw[-stealth](lastRNN.north)--(avg.south east)node[midway,above]{$o_T$};

\node[draw,
minimum width=1cm,
    minimum height=1cm,
    align=center
](FF) at (-1.75,6){Feed forward layer};

\node[draw,
minimum width=1cm,
    minimum height=1cm,
    align=center
](xhat) at (-1.75,8){$\hat{X}$};

\draw[-stealth](avg.north)--(FF.south){};
\draw[-stealth](FF.north)--(xhat.south)node[midway,left]{softmax()};
\end{tikzpicture}
\caption{\textbf{A high level overview of   RNNInference.} Each action observation pair $(a_t,y_t)$ passes through a RNN architecture and outputs $o_t$. All outputs are averaged and passed through a feed forward layer. Finally a softmax activation function produces an estimate of the true hypothesis. A similar architecture is employed by RNNMonitor, without the softmax function.  } \label{fig:M2}
\end{figure}
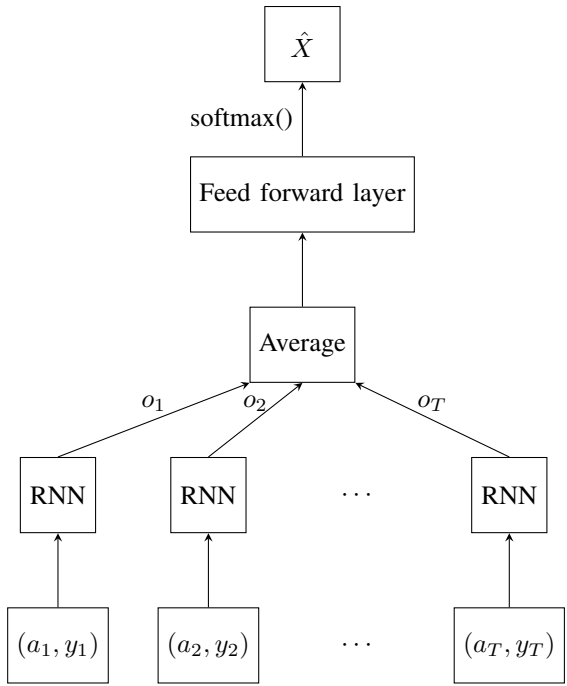

Each network is fed by the sequence of action and observation pairs. At each time  $t$, a tuple $(a_t,y_t)$ passes through the RNN cell and an output $o_t$ is generated. The mean of all outputs goes through a feed forward layer that outputs the prediction. In the case of classification, the feed forward layer is followed by a softmax activation function. 
The loss function of RNNinference is the cross entropy loss between the true hypothesis $X$ and the predicted hypothesis. The loss function of RNNmonitor is the mean squared error loss.
All networks  are developed using the pytorch framework \cite{pytorchCite}. A high level overview of RNNinference is provided at fig \ref{fig:M2}.

Training of  the  neural networks takes place in  three phases. First, we train the RNNpolicy network using reinforcement learning on an artificial simulation environment. Then the RNNmonitor and subsequently the RNNinference are trained using the adam optimizer \cite{adam}. In the fixed horizon setting, training of the RNNmonitor is deactivated. 

The RNNpolicy is initialised with random parameters. The artificial simulation environment is created as follows. At each training episode the underlying hypothesis $i$ is sampled from $\mathcal{X}$ according to $\rho_0$.
At each time $t$ the agent chooses an action $a_t$ based on the state of the recurrent neural network and its most recent observation $y_{t-1}$ (action $a_1$ is chosen randomly). An observation $y_t$ is sampled from $\mathcal{Y}$ according to $P[Y|X=i,a=a_t]$. The training environment internally updates the belief vector $\rho_{t}$.
The reward provided to the agent is the error improvement  $\gamma_{t-1}-\gamma_{t}$. After $T$ time steps the training episode terminates. In the finite horizon setting, $T$ defines the horizon length. In the infinite horizon setting, we set $T$ to a large value to ensure sufficient exploration.   We use a reliable open source implementation of recurrent PPO \cite{stable-baselines3} (contributed version).

Upon completion of phase 1 and training of the  RNNpolicy network, a new dataset is generated by randomly selecting the horizon of each episode. Each data point consists of a sequence of actions and observations labeled by the error probability at the final time step. The resulting dataset is used to train RNNmonitor in stage 2. 

Finally training of RNNinference is realized via a new dataset generated as follows. At each time instance $t$ of each training episode, RNNmonitor outputs an approximation of the error probability $\bar{\gamma}_t$. If $\bar{\gamma}_t<c$, where the hyperparameter $c$ is a user defined threshold, the episode terminates, else RNNpolicy continues with the experiment selection. The label of each data point is the underlying true hypothesis. The desired dataset is obtained by repeating this procedure for multiple episodes. 

In the execution phase, the three networks are run and evaluated in a testing environment which slightly differs from the one they are trained on.

\section{Case study: Anomaly detection}
We consider an anomaly detection example involving two sensors $\mathcal{A}$ and $\mathcal{B}$ that detect anomalies in their proximity. There are 4 possible hypotheses. $X=0$ means there is no anomaly (the system is in safe state). $X=1$ and $X=2$ mean there is an anomaly near sensors $\mathcal{A}$ and $\mathcal{B}$ respectively. Finally, $X=3$ means there are anomalies near both sensors. We assume a uniform prior. We create two slightly different environments, one for training the DRL agent and the RNN decoders and one for testing their performance. The observation models are summarized in tables  \ref{tab:obs1} and \ref{tab:obs2}. 
\begin{table}
    \centering
    \caption{$P[Y=1|a_t,X]$ under training}
    \label{tab:obs1}
    \begin{tabular}{|c|c|c|c|c|}
    \hline
         &$X=0$ & $X=1$ & $X=2$& $X=3$  \\
    \hline
         $\mathcal{A}$&0.2&0.8&0.2&0.8\\
         \hline
         $\mathcal{B}$&0.2&0.2&0.8&0.8\\
         \hline
    \end{tabular}
    
\end{table}
\begin{table}
    \centering
    \caption{$P[Y=1|a_t,X]$ under testing}
    \label{tab:obs2}
    \begin{tabular}{|c|c|c|c|c|}
    \hline
         &$X=0$ & $X=1$ & $X=2$& $X=3$  \\
    \hline
         $\mathcal{A}$&0.25&0.75&0.25&0.75\\
         \hline
         $\mathcal{B}$&0.15&0.15&0.85&0.85\\
         \hline
    \end{tabular}
    
\end{table}

First we examine how efficiently LSTMs and GRUs learn to map action and observation sequences to $\gamma_t$, $LL_t$, $C_t$ and $\hat{i}_t$, where $\hat{i}_t$ is the most likely hypothesis. 
We follow the modified Chernoff strategy in the training environment as described in \cite{controlledSensForMH}, and collect a dataset of 60000 sequences. The first 50000 are used for training and the rest for validation. We also build a test set of 10000 sequences using the testing environments.  We pick a fixed horizon at the beginning of each episode, randomly in the range 5-50. 

We consider unidirectional and  bidirectional  RNNs with two hidden layers and  size in the range 50-400. The RNN with the best score on the validation set is used in the test set. In case of ties, the simpler architecture is preferred. The results for all 5 metrics are shown in table 
\ref{tab:supervRes}.  

We note that as the error probability becomes small, both $LL_t$ and $C_t$ can be large. Moreover, they might take small or even negative values for small horizons. Hence we discarded all samples where $LL_t>100$, because the maximum absolute error (MAE) was very large. On the other hand $\gamma_t$ is always in the range $[0,1]$. It can be seen from table \ref{tab:supervRes} that both LSTMs and  GRUs learn $\hat{j}_t$ and $\gamma_t$ very accurately. For this reason, the stopping rule in the infinite horizon setting employs the error probability.
\begin{table}
\caption{Test scores and size parameters.}
    \label{tab:supervRes}
    \centering
    \begin{tabular}{|c|c|c|}
    \hline
         & LSTM &GRU  \\
    \hline
         best hidden size for $LL_t$& 400(BI)&300 (BI) \\
         \hline
        best hidden size for $\hat{j}_t$&200 &200 \\
\hline
best hidden size for $C(\rho_t)$ &250(BI) &250(BI) \\
\hline
best hidden size for $\gamma_t$& 200&250 \\
\hline
    precision on $\hat{j}_t$&0.9996 & 0.9967\\
    \hline
    recall on $\hat{j}_t$&0.9996 & 0.996\\
    \hline
    f1 score on $\hat{j}_t$&0.9996 &0.9996 \\
    \hline
     MAE for $LL_t$& 13.25& 14.397\\
     \hline
     MAE for $C(\rho_t)$& 0.952& 0.945\\
     \hline
     MAE for $\gamma_t$&0.00085 &0.0019 \\
     \hline
         
    \end{tabular}
    
\end{table}

We explore different horizons of length 10, 25, 50, and 100. For each horizon, the PPO agent is trained and a dataset that has 60000 different sequences of actions and observations is generated.
Both decoders are unidirectional RNNs with 2 hidden layers of the same size. The GRU decoder has 250 hidden units and the LSTM decoder 200 units.
When paired with a GRU or an LSTM decoder, for horizons 10-50, PPO's performance is comparable to that of the Chernoff test \footnote{Note that the Chernoff test has an advantage because it knows the actual error probabilities of the testing environment. It can not be deployed in the belief free setting. In contrast, our model has only seen data from the training environment.}. In fact, for $T=10$ it achieves a smaller error probability. For $T=100$, the Chernoff test achieves a significantly smaller error probability. The GRU decoder  performs slightly better than the LSTM decoder.
\begin{table}
\caption{Average error probability over 10000 episodes}
    \label{tab:resFixedH}
    \centering
    \begin{tabular}{|c|c|c|c|}
    \hline
         T&PPO-LSTM&PPO-GRU&Chernoff  \\
    \hline
        10 &0.152 & 0.151&0.162 \\
        \hline
        25 &0.0396 &0.0341
 & 0.0312\\
        \hline
        50&0.0078&0.008& 0.0017\\
        \hline 
        100&0.0008&0.0007&0.00005 \\
        \hline
    \end{tabular}
    
\end{table}

In the infinite horizon case, we use the PPO agent  previously trained for $T=50$. The dataset consists of 60000 training examples. The horizon of each episode is chosen uniformly in the range [1,50]. The LSTM network approximates the error probability. When the network detects that the error probability is below a user defined threshold $c$, the episode terminates. Using the agent and the network a new dataset of 60000 examples is generated and used to train another LSTM that infers the underlying hypothesis. We tested the PPO agent based on the LSTM networks in the test environments for 10000 episodes. We repeated the procedure for GRU networks.
The test results for different tolerance levels are demonstrated in  table \ref{tab:resInfH}.

The LSTM networks are bidirectional with 2 hidden layers of 200 neurons. They also employ  dropout with probability $p=0.2$. The GRUs are bidirectional with 4 hidden layers of 250 neurons. Again, dropout is used with $p=0.2$. 

Contrary to the finite horizon, LSTM decoders perform  better than GRU decoders. When the PPO agent is combined with LSTM decoders, it terminates at the desired error probability level slightly faster than the Chernoff test in 2/4 experiments.

\begin{table}
\centering
\caption{Average stopping time and error probability} \label{tab:resInfH}
\begin{subtable}[b]{0.3\textwidth}
\caption{stopping time}
\begin{tabular}{|c|c|c|c|}
\hline
   c&PPO-LSTM & PPO-GRU&Chernoff    \\
   \hline
   0.3 & 5.85  & 6 & 4.58\\
   \hline
   0.2 &7.41 &8 &7.73 \\
   \hline
   0.1 &11.58 & 13&12.61 \\
   \hline
   0.05 &12.66& 16.14 &12.164\\
   \hline
   \end{tabular}
\end{subtable}\\

\begin{subtable}[b]{0.3\textwidth}
\caption{error probability}
\begin{tabular}{|c|c|c|c|}
\hline
c&PPO-LSTM & PPO-GRU&Chernoff    \\
   \hline
   0.3 & 0.198  & 0.205 & 0.248\\
   \hline
   0.2 & 0.134& 0.164 & 0.131\\
   \hline
   0.1 &0.075 &0.092 & 0.046\\
   \hline
   0.05 & 0.043 &0.051 & 0.038\\
   \hline
   \end{tabular}
\end{subtable}
\end{table}
\section{Further examples}
\label{sec:largeEXAMPLE}
Next we  consider a larger anomaly detection problem. Four sensors monitor four independent random processes. Any number of processes can be abnormal, therefore there are 16 different hypotheses. Proceeding as in the previous chapter, we  build two slightly different training and testing environments. 
When a process is abnormal in the training environment the sensor outputs 1 with probability 0.8 and 0 with probability 0.2. When the process is normal the numbers are reversed. In the testing environment, the first two sensors output 1 with probability 0.85 when the processes are  abnormal, and the last two output 1 with probability 0.75. 

We repeat the fixed horizon experiments, only this time, the training dataset consists of 150000 examples. The network pairs are tested for 10000 episodes. The results  for different horizons are summarized in table  \ref{tab:resLarge}.
\begin{table}[]
\caption{Average error probability in the four sensors case}
    \label{tab:resLarge}
    \centering
    \begin{tabular}{|c|c|c|c|}
    \hline
         horizons& PPO+LSTM & PPO+GRU & Chernoff \\
        \hline
        10 & 0.48 &0.49 &0.53\\
        \hline
        25&0.247 & 0.245 & 0.234\\
        \hline
        50&0.0698&0.0699&0.071\\
        \hline
        100&0.0238 &0.0157&0.008\\
        \hline
    \end{tabular}
    
\end{table}
Both network pairs perform better than the Chernoff test for horizons of length 10 and 50. For horizons 25 and 100, the Chernoff test performs slightly better. 

Next we consider  an anomaly detection example that lies closer to real world data, namely the windows 10 dataset from the TON\_IOT cybersecurity data  \cite{moustafa2020federated}.  More information about how the environment works is provided in the extended version of this paper available at arxiv.

We repeat the experiments of table \ref{tab:resLarge}. The results are shown in table \ref{tab:resLarge2}. Our method performs better than the Chernoff test for horizons $T=10$ and $T=25$, and the difference for $T=50$ and $T=100$ is very small (one error out of 10000 episodes).

\begin{table}[]
\caption{Average error probability out of 10000 episodes for the cybersecurity data.}
    \label{tab:resLarge2}
    \centering
    \begin{tabular}{|c|c|c|c|}
    \hline
         horizons& PPO+LSTM & PPO+GRU & Chernoff \\
        \hline
        10 &0.0632  &0.0634
 &0.2314\\
        \hline
        25&0.0064 &0.0057 &0.0194 \\
        \hline
        50&0001&0.0002&0.0001\\
        \hline
        100&0.001 &0&0\\
        \hline
    \end{tabular}
    
\end{table}

The sample efficiency of the supervised decoders and the recurrent PPO algorithm are analyzed in the extended version. The DRL algorithm requires far more samples than the supervised decoders.

\section{Conclusion} 
We have demonstrated that model free recurrent DRL combined with RNN decoders can successfully be applied to ASHT environments. Unlike other approaches, our method does not need to recursively construct large belief or likelihood vectors, hence it can  scale up to very large environments. We have also employed different training and testing environments. The proposed method  performs competitively with the asymptotically optimal  Chernoff test, both in finite and infinite horizon problems, despite the lack of knowledge of the observation probabilities. For smaller horizons it usually performs better. 

In future work we will seek to improve the sample complexity of the DRL part of our method, and  apply offline (batch) reinforcement learning for large environments when there is not an accurate simulator and only a fixed dataset. We will consider replacing LSTMs and GRUs with transformers \cite{attentionAllYouNeed}.

\bibliographystyle{ieeetr} 
\bibliography{references}

\section{Extended version: Testing and Training on cybersecurity data}

Each entry of the dataset contains information about one process running in a windows 10 PC. It provides information on the processes characteristics such as the activity of processor, network, memory and disk. Each process has a binary label about its status (normal or abnormal). we ignore the second label describing the type of attack. The following five features are chosen:
\begin{enumerate}
    \item 'Memory Demand Zero Faults sec' ,
    \item 'Network\_I(Intel R \_82574L\_GNC)Bytes Received sec',
    \item 'Network\_I(Intel R \_82574L\_GNC) Bytes Sent sec',
    \item'Process\_Page Faults\_sec',
    \item 'Memory Page Writes sec',
\end{enumerate}
we used the default sklearn implementation of a decision tree \cite{scikit-learn}, setting class\_weight='balanced', combined with a standard scaler, to perform classification.

We split the windows 10 dataset in three sets a training set, a validation set, and a test set. The model is trained on the training set, and the conditional observation probabilities are estimated on the validation set. They are then used to train the algorithm.

The testing environment works as follows. First, the true hypothesis is sampled from the uniform prior. Then, at each time, depending on whether the queried process is normal or not we sample an entry from the test set and pass it from the trained decision tree classifier. The  output of the classifier ($0$ or $1$) is used to update the belief. The other features of the environment remain unchanged.  

\subsection{Sample efficiency}
\begin{figure}[]
    \centering
    \caption{Error probability of the GRU decoder for different dataset sizes}
    \scalebox{0.85}{\includegraphics{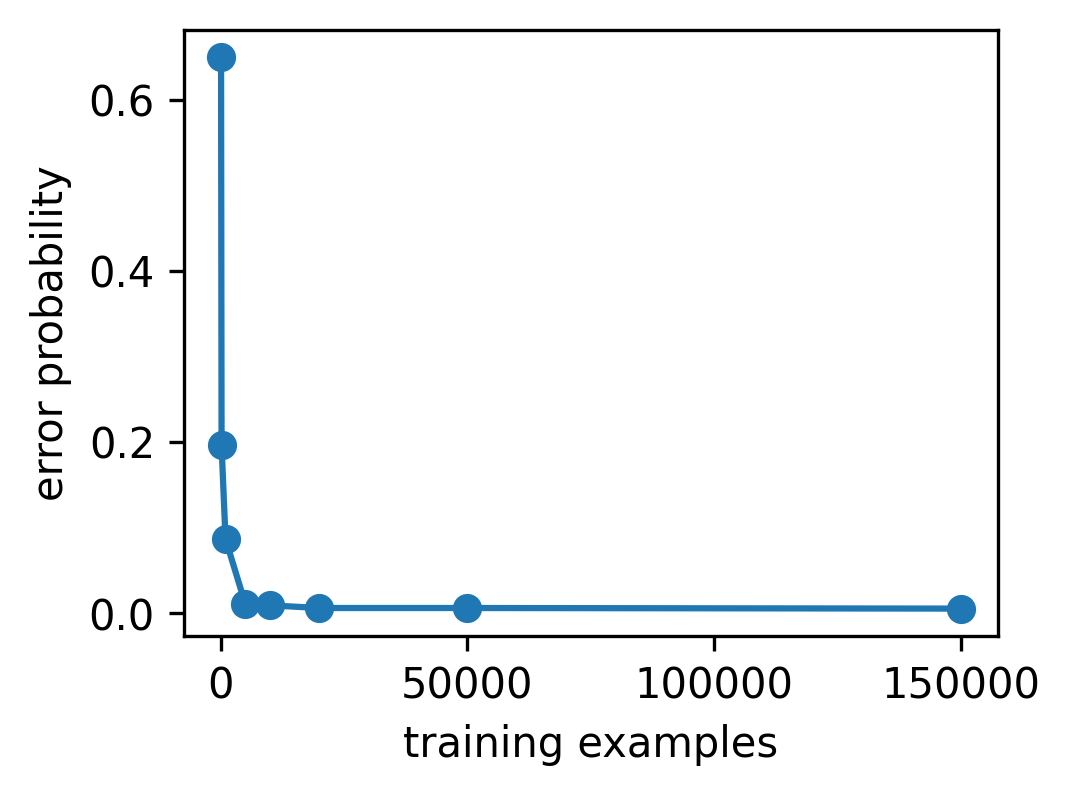}}
    \label{fig:gruvsize}
\end{figure}

\begin{figure}[]
    \centering
    \caption{Error probability of the LSTM decoder for different dataset sizes}
    \scalebox{0.85}{\includegraphics{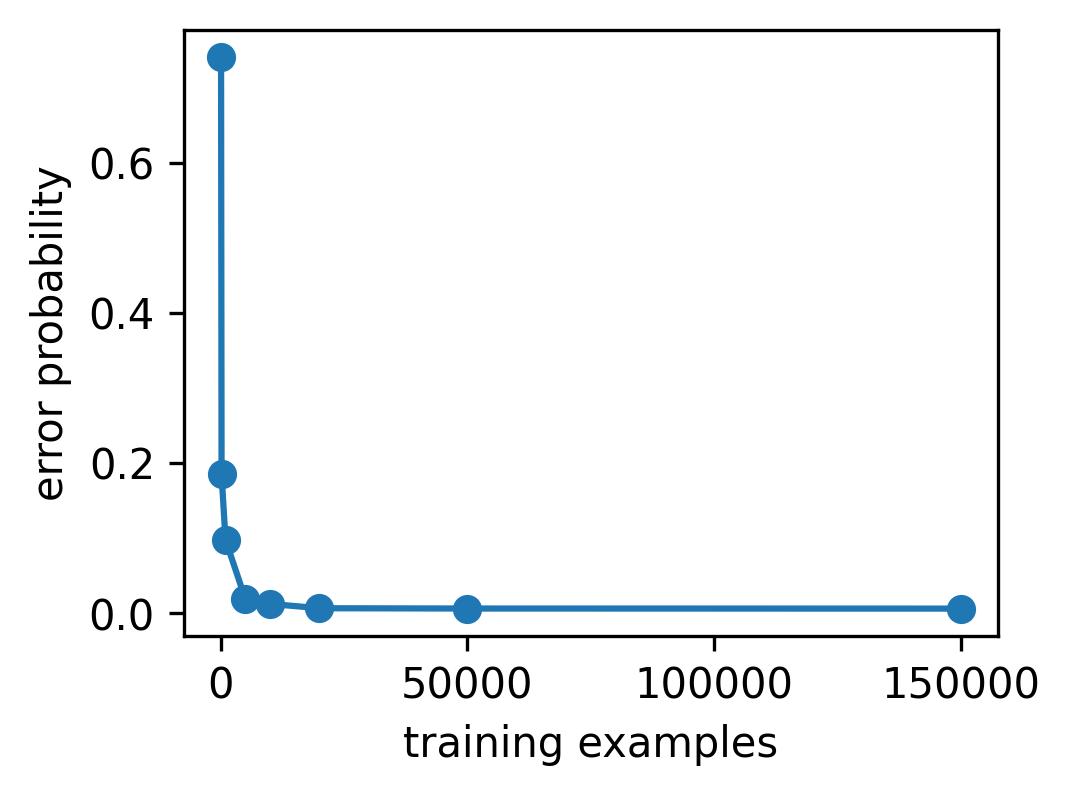}}
    \label{fig:lstmvsize}
\end{figure}

We set $T=25$ and collect datasets of different sizes from $100$ to $150000$ to train the decoders using an already trained PPO-LSTM agent. The testing remains the same. The error probabilities are illustrated in Figures \ref{fig:gruvsize} and \ref{fig:lstmvsize} respectively. We observe that after $1000$ training examples the error probability becomes very small, and after $20000$ it is near zero and does not decrease significantly any further. Results for other horizons are similar.

\begin{figure*}
     \centering
     \begin{subfigure}[t]{0.45\textwidth}
         \centering         \includegraphics[width=\textwidth]{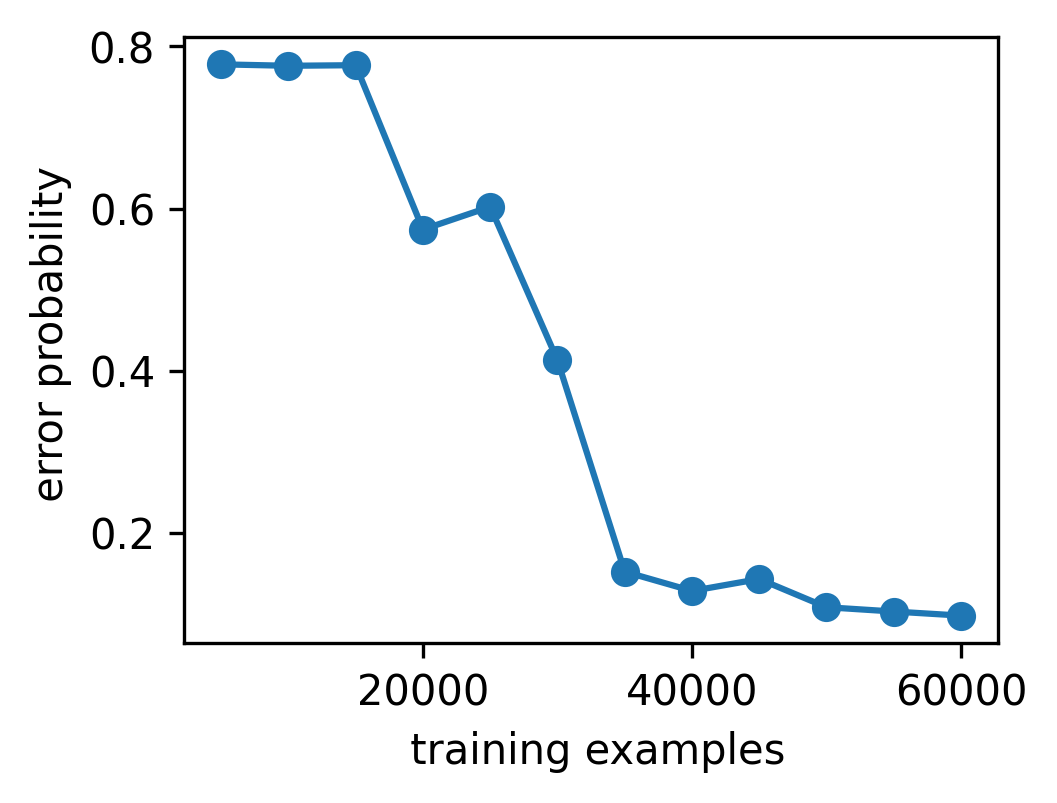}
         \caption{T=10}
     \end{subfigure}
     \hfill
     \begin{subfigure}[t]{0.45\textwidth}
         \centering
    \includegraphics[width=\textwidth]{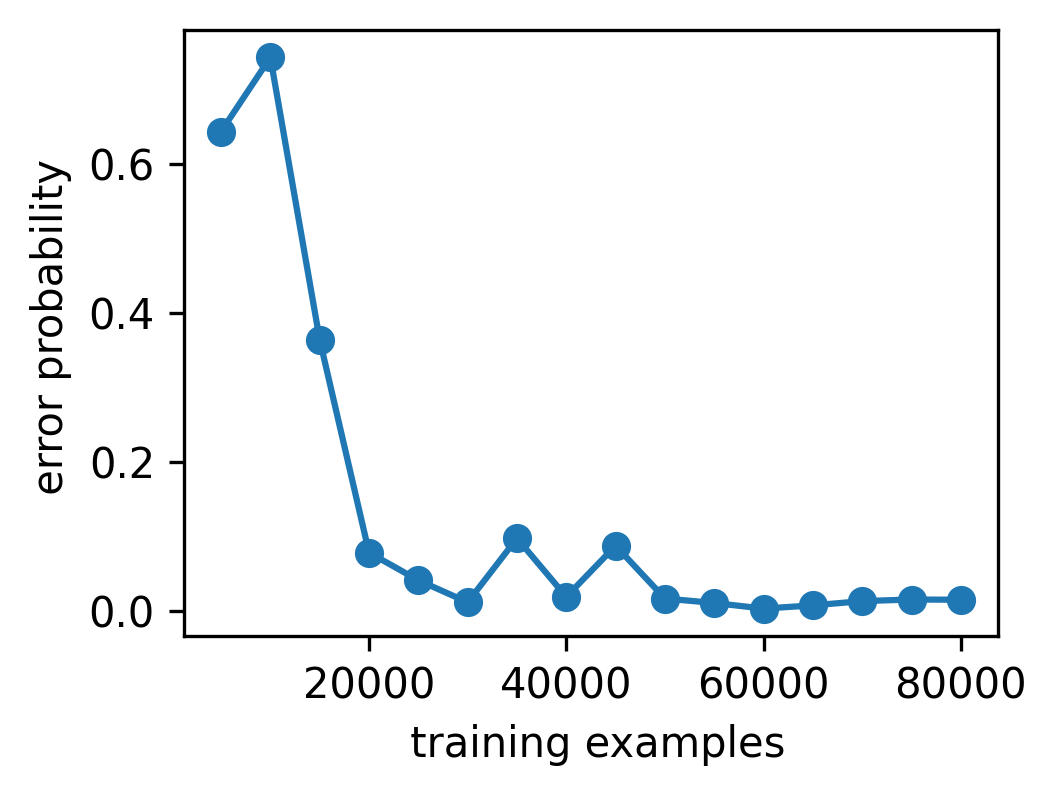}
         \caption{T=25}
     \end{subfigure}
     
     \begin{subfigure}[b]{0.45\textwidth}
         \centering
    \includegraphics[width=\textwidth]{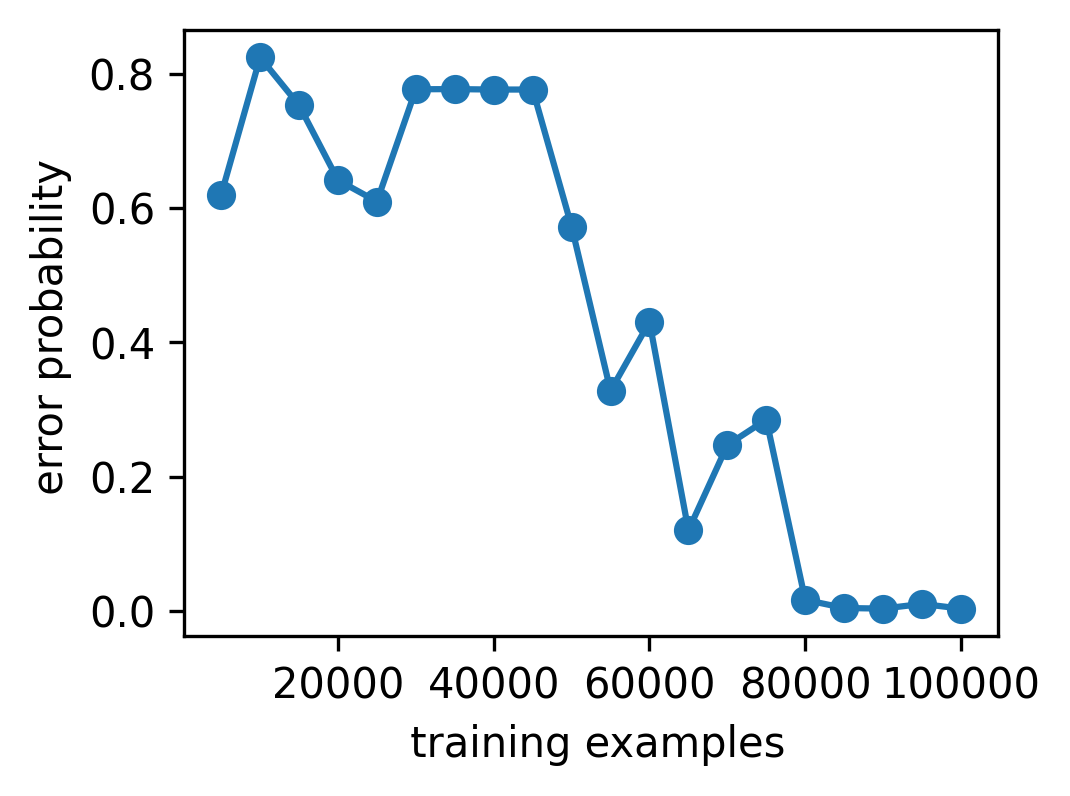}
         \caption{T=50}
     \end{subfigure}
     \begin{subfigure}[b]{0.45\textwidth}
         \centering
    \includegraphics[width=\textwidth]{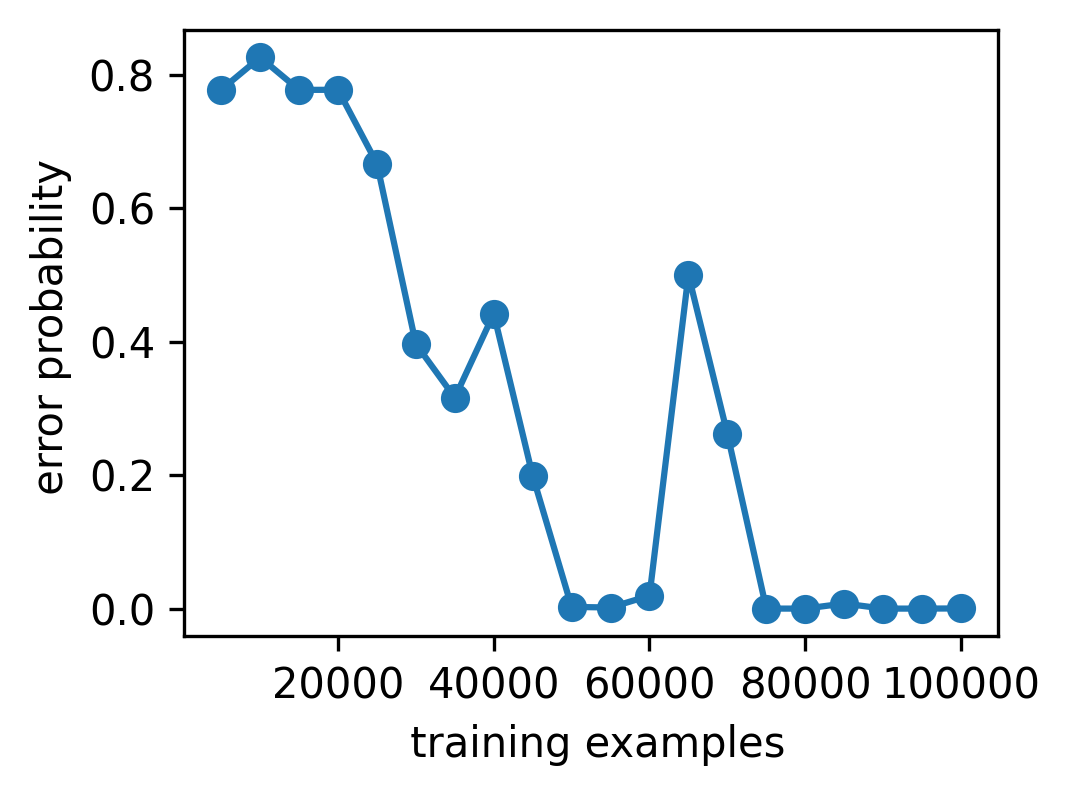}
         \caption{T=100}
     \end{subfigure}
        \caption{Sample efficiency of PPO-LSTM for different horizons}
        \label{fig:PPOSAMPLE}
\end{figure*}

Next, we evaluate the sample efficiency of the PPO-LSTM algorithm. We trained the model for a large number of episodes and interrupted for testing every $5000$ episodes. For testing we  use the MAP decoder.  The results  are shown in figure \ref{fig:PPOSAMPLE}.

We observe that the algorithm is more sample efficient for smaller horizons and the  decoders need far fewer samples than the DRL algorithm.

\end{document}